\documentclass[10pt,twocolumn,letterpaper]{article}

\usepackage[pagenumbers]{cvpr} %

\usepackage{graphicx}
\usepackage{amsmath}
\usepackage{amssymb}
\usepackage[utf8]{inputenc} %
\usepackage[T1]{fontenc}    %
\usepackage{url}            %
\usepackage{booktabs}       %
\usepackage{amsfonts}       %
\usepackage{nicefrac}       %
\usepackage{microtype}      %
\usepackage{xcolor}         %
\usepackage{multirow}
\usepackage{graphicx}
\usepackage{makecell}
\usepackage{amsmath,amssymb}
\usepackage{overpic}
\usepackage{adjustbox}
\usepackage{soul}
\makeatletter
\@namedef{ver@everyshi.sty}{}
\makeatother
\usepackage{tikz}
\usetikzlibrary{arrows.meta}

\newcommand{\Best}[1]{\textbf{\textcolor{black}{#1}}}
\newcommand{\B}[1]{\Best{#1}}

\newcommand{\myparagraph}[1]{\medskip\noindent{\bf#1}\,}

\usepackage[pagebackref,breaklinks,colorlinks]{hyperref}

\usepackage[capitalize]{cleveref}
\crefname{section}{Sec.}{Secs.}
\Crefname{section}{Section}{Sections}
\Crefname{table}{Table}{Tables}
\crefname{table}{Tab.}{Tabs.}

\def\cvprPaperID{} %
\def\confName{}
\def\confYear{}

\begin{document}

\title{Regularization of NeRFs using differential geometry}

\author{%
  Thibaud~Ehret \quad
  Roger~Mar\'i \quad
  Gabriele~Facciolo \\
  Université Paris-Saclay, CNRS, ENS Paris-Saclay, Centre Borelli\\
  91190, Gif-sur-Yvette, France\\
{\tt\small thibaud.ehret@ens-paris-saclay.fr}
}
\maketitle

\begin{abstract}
Neural radiance fields, or NeRF, represent a breakthrough in the field of novel view synthesis and 3D modeling of complex scenes from multi-view image collections. Numerous recent works have shown the importance of making NeRF models more robust, by means of regularization, in order to train with possibly inconsistent and/or very sparse data. In this work, we explore how differential geometry can provide elegant regularization tools for robustly training NeRF-like models, which are modified so as to represent continuous and infinitely differentiable functions. In particular, we present a generic framework for regularizing different types of NeRFs observations to improve the performance in challenging conditions. We also show how the same formalism can also be used to natively encourage the regularity of surfaces by means of Gaussian or mean curvatures.
\end{abstract}

\section{Introduction}

Realistic rendering of new views of a 3D scene or a given volume is a long standing problem in computer graphics. The interest in this problem has been rekindled by the growth of augmented and virtual reality. Traditionally, 3D scenes were estimated from a set of images using classic Structure-from-Motion (SfM) and Multi-View Stereo (MVS) tools such as COLMAP~\cite{schonberger2016structure} or \cite{snavely2006photo,moulon2016openmvg,rupnik2017micmac,furukawa2007accurate}.

\begin{figure}
    \hspace{-0.3cm}
    \begin{tabular}{c@{\hskip 0.0\linewidth}c@{\hskip 0.0\linewidth}c}
    \rotatebox[origin=c, x=0cm]{90}{\scriptsize DiffNeRF (depth)} \includegraphics[width=0.32\linewidth]{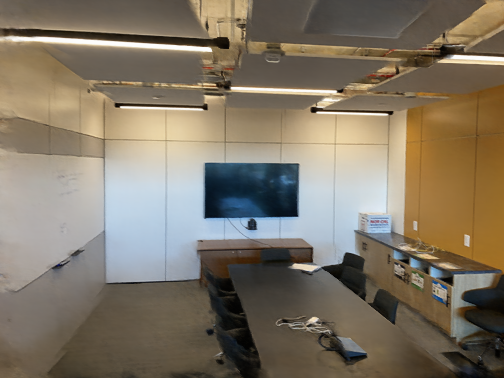} &
    \includegraphics[width=0.32\linewidth]{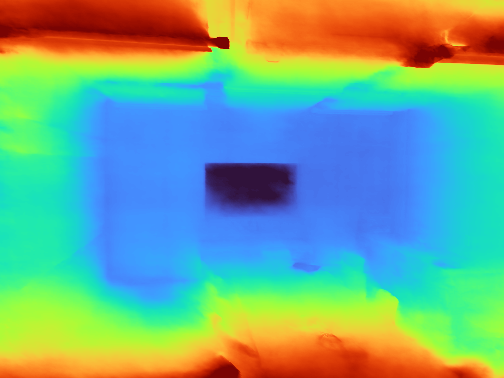} &
    \includegraphics[width=0.32\linewidth]{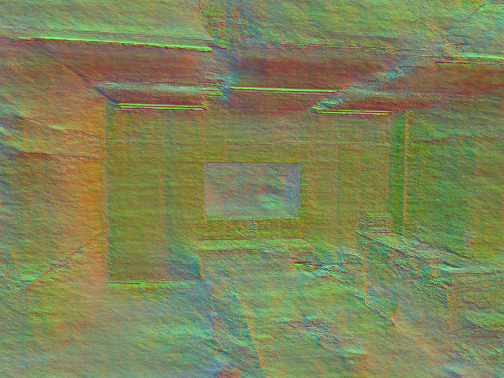} \\
    \rotatebox[origin=c, x=.1cm]{90}{\scriptsize DiffNeRF (normals)} \includegraphics[width=0.32\linewidth]{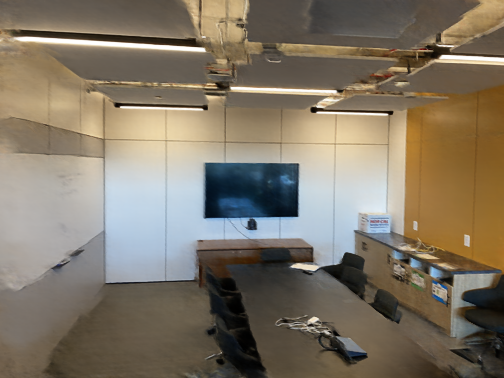} &
    \includegraphics[width=0.32\linewidth]{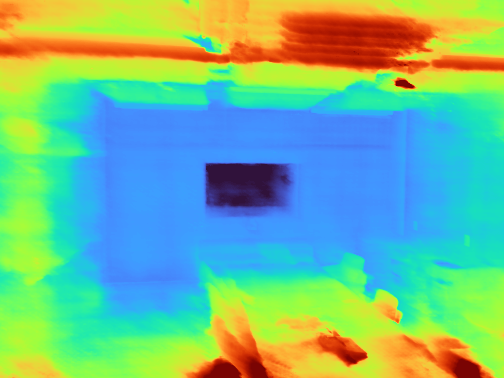} &
    \includegraphics[width=0.32\linewidth]{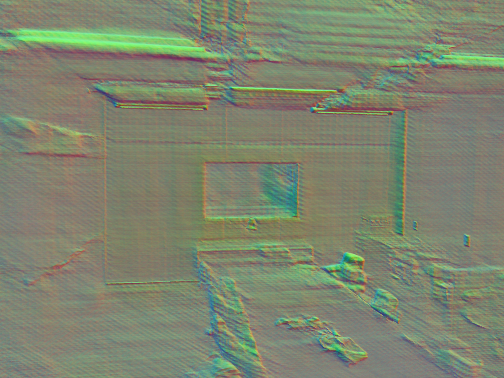} \\
    \rotatebox[origin=c, x=-0.1cm]{90}{\scriptsize RegNeRF~\cite{niemeyer2021Regnerf}} \includegraphics[width=0.32\linewidth]{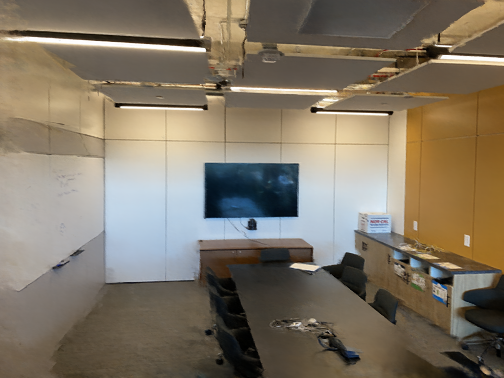} &
    \includegraphics[width=0.32\linewidth]{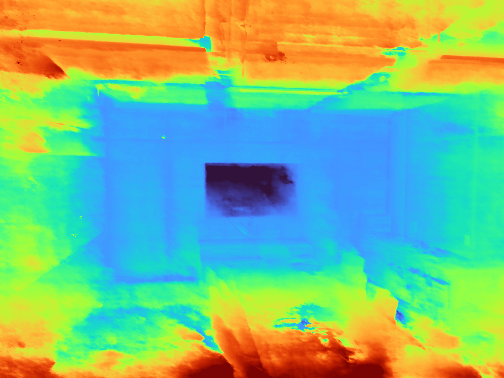} &
    \includegraphics[width=0.32\linewidth]{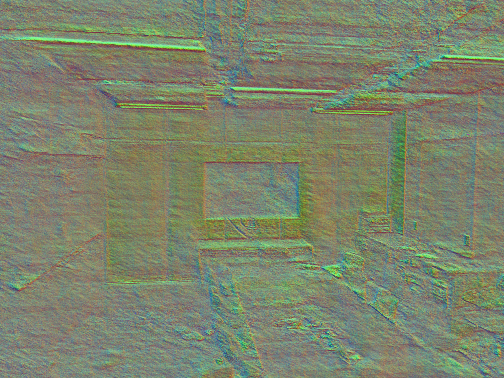} \\
    \end{tabular}
    \vspace{-.5em}
    \caption{We propose a generic regularization framework for NeRF that outperforms previous state-of-the-art methods when training with \ul{only three input views}. We compare here the proposed DiffNeRF with depth regularization (top), DiffNeRF with normals regularization (middle) and RegNeRF~\cite{niemeyer2021Regnerf}. Left to right: RGB prediction, depth map, map of normals.}
    \label{fig:teaser}
\end{figure}

Recently, Mildenhall~\etal~\cite{mildenhall2020nerf} have shown that differentiable volume rendering operations can be plugged into a neural network to learn a neural radiance field (NeRF) volumetric representation of a scene encoding its geometry and appearance. Starting from a sparse, yet nonetheless large, set of views of the scene, NeRF learns in a self-supervised manner, by maximizing the photo-consistency across the predicted renderings corresponding to the available viewpoints. After convergence, the network is able to render realistic novel views by querying the NeRF function at unseen viewpoints.

This breakthrough led to a very active research field focused on pushing back the limits of the initial models. Notably, Martin-Brualla~\etal~\cite{martin2021nerf} showed that it is possible to learn scenes from unconstrained photos with moving transient objects or different lightning conditions. Other works deal with dynamic or deformable scenes~\cite{pumarola2021d, tretschk2021nonrigid, park2021nerfies, park2021hypernerf, li2021neural}, complex illumination models~\cite{srinivasan2021nerv, boss2021nerd, zhang2021nerfactor, mari2022sat} or very few training views~\cite{yu2021pixelnerf,jain2021putting, kim2021infonerf}. In other words, the goal is to make NeRF more robust to be able to train reliably even in the most adverse conditions. For example, imposing regularity constraints on the scene seems to be a promising way to reduce reliance on large datasets~\cite{niemeyer2021Regnerf}.

{The objective of this work is to show how one can adapt differential geometry concepts to elegantly incorporate regularizers that make NeRF more robust}. The advantages are twofold: first, differential geometry is mathematically formalized and its literature is vast with many suitable tools already available and, second, neural representations are perfectly adapted to represent continuous infinitely differentiable volumetric functions in which differential geometry operators are naturally defined.

To this aim, we present a generic framework based on differential geometry to regularize different types of NeRFs observations. We derive the two specific cases of depth regularization, thus linking to the previously proposed RegNeRF~\cite{niemeyer2021Regnerf}, as well as normals regularization in Section~\ref{sec:regnerf}. We also show in Section~\ref{sec:exps} that this approach is not only competitive with the state of the art for the problem training a NeRF model when few images (for example only three) are available but also that it produces smoother and more accurate depth maps.
Finally, we straightforwardly extends the proposed framework to surfaces regularization in Section~\ref{sec:surface_reg} showing that generality of the proposed approach.

\section{Related Work}

\subsection{Fundamentals of Neural Radiance Fields}
\label{sec:pres_nerf}

NeRF was originally introduced as a continuous volumetric function $\mathcal{F}$, learned by a multi-layer perceptron (MLP), to model the geometry and appearance of a 3D scene~\cite{mildenhall2020nerf, tewari2020state}. Given a 3D point $\mathbf{x} \in \mathbb{R}^3$ of the scene and a viewing direction $\mathbf{v} \in \mathbb{R}^2$, NeRF predicts an associated RGB color~$\mathbf{c} \in [0, 1]^3$ and a scalar volume density~$\sigma \in [0, \infty)$, \ie
\begin{equation}
    \mathcal{F}:(\mathbf{x}, \mathbf{v}) \mapsto (\mathbf{c}, \sigma).
    \label{eq:classic_nerf_inputs_outputs}
\end{equation}
The value of $\sigma$ defines the geometry of the scene and is learned exclusively from the spatial coordinates $\mathbf{x}$, while the value of $\mathbf{c}$ is also dependent on the viewing direction $\mathbf{d}$, which allows to recreate non-Lambertian surface reflectance.

NeRF models are trained based on a classic differentiable volume rendering operation~\cite{max1995optical}, which establishes the resulting color of any ray passing through the scene volume and projected onto a camera system. Each ray $\mathbf{r}(t) = \mathbf{o} + t\mathbf{v}$ with $t\in \mathbb{R}^+$, defined by a point of origin $\mathbf{o}$ and a direction vector $\mathbf{v}$, is discretized into $N$ 3D points $\{\mathbf{x}_i\}_{i=1}^N$, where $\mathbf{x}_i = \mathbf{o} + t_i\mathbf{v}$ with $t_i$ sampled between the minimum and maximum depth. The sampling depends on the method. The rendered color $\mathbf{c}(\mathbf{r})$ of a ray $\mathbf{r}$ is obtained as
\begin{equation}
    \mathbf{c}(\mathbf{r}) = \sum_{i=1}^{N}T_i\alpha_i\mathbf{c}_i
    \label{eq:nerf_color_rendering}\vspace{-1em}
\end{equation}
where
\begin{equation}
    T_i = \prod_{j=1}^{i-1} \left( 1 - \alpha_j \right)\hspace{0.5em} \text{and}\hspace{0.5em} \alpha_i = 1 - \exp(-\sigma_i(t_{i+1}-t_{i})). 
\end{equation}

In \eqref{eq:nerf_color_rendering}, $\mathbf{c}_i$ and $\sigma_i$ correspond to the color and volume density output by the MLP at the $i$-th point of the ray, \ie $\mathcal{F}(\mathbf{x}_i, \mathbf{v})$ as per \eqref{eq:classic_nerf_inputs_outputs}. The transmittance $T_i$ and opacity $\alpha_i$ are two factors that weight the contribution of the $i$-th point to the rendered color according to the geometry described by $\sigma_i$ and $\sigma_j : j < i$, to handle occlusions.

Using the transmittance $T_i$ and opacity $\alpha_i$, the observed depth $d(\mathbf{r})$ in the direction of a ray $\mathbf{r}$ can be rendered in a similar manner to \eqref{eq:nerf_color_rendering} \cite{deng2021depth, roessle2021dense} as
\begin{equation}
    d(\mathbf{r}) = \sum_{i=1}^{N}T_i\alpha_it_i.
    \label{eq:depth_nerf}
\end{equation}
Following this volume rendering logic, the NeRF function $\mathcal{F}$ is optimized by minimizing the squared error between the rendered color and the real colors of a batch of rays $\mathcal{R}$ that project onto a set of training views of the scene taken from different viewpoints:
\begin{equation}
    L_{RGB} = \sum_{\mathbf{r} \in \mathcal{R}} \|  \mathbf{c}(\mathbf{r}) - \mathbf{c}_{\text{GT}}(\mathbf{r}) \|_2^2,
    \label{eq:nerf_classic_loss} 
\end{equation}
where $\mathbf{c}_{\text{GT}}(\mathbf{r})$ is the observed color of the pixel intersected by the ray $\mathbf{r}$, and $\mathbf{c}(\mathbf{r})$ is the NeRF prediction \eqref{eq:nerf_color_rendering}. The rays in each batch $\mathcal{R}$ are chosen randomly from the available views, encouraging gradient flow where rays casted from different viewpoints intersect with consistent scene content.

\myparagraph{mip-NeRF:} The original NeRF approach casts a single ray per pixel \cite{mildenhall2020nerf}. When the training images observe the scene at different resolutions, this  can lead to blurred or aliased renderings. To prevent such situation, the mip-NeRF formulation~\cite{barron2021mipnerf} can be adopted, which casts a cone per pixel instead. As a result, mip-NeRF is queried in terms of conical frustums and not discrete points, yielding a continuous and natively multiscale representation of regions of the volume space.

\subsection{Regularization in Few-shot Neural Rendering}

The original NeRF methodology was demonstrated using 20 to 62 views for real world static scenes, and 100 views or more for synthetic static scenes \cite{mildenhall2020nerf}. In the absence of large datasets, the MLP usually overfits to each training image when only a few are available, resulting in unrealistic interpolation for novel view synthesis and poor geometry estimates.

A number of NeRF variants have been recently proposed to address few-shot neural rendering. The use of regularization techniques is common in these variants, to achieve smoother results in unobserved areas of the scene volume or radiometrically inconsistent observations \cite{kim2021infonerf}. 

\myparagraph{Implicit/indirect regularization} methods rely on geometry and appearance priors learned by pre-trained models. PixelNeRF \cite{yu2021pixelnerf} introduced a framework that can be trained across multiple scenes, thus acquiring the ability to generalize to unseen environments. The MLP learns generic operations while keeping the output conditioned to scene-specific content thanks to an additional input feature vector, extracted by a pre-trained convolutional neural network (CNN). Similarly, \mbox{DietNeRF}~\cite{jain2021putting} complements the NeRF loss~\eqref{eq:nerf_classic_loss} with a secondary term that encourages similarity between pre-trained CNN high-level features in renderings of known and unknown viewpoints. Other approaches like GRAF~\cite{schwarz2020graf}, $\pi$-GAN \cite{chan2021pi}, Pix2NeRF \cite{cai2022pix2nerf} or LOLNeRF\cite{rebain2021lolnerf} combine NeRF with generative models: latent codes are mapped to an instance of a radiance field of a certain pre-learned category (e.g. faces, cars), thus providing a reasonable 3D model regardless of the number of available of views.

\myparagraph{Explicit/direct regularization} methods can be divided into externally supervised and self-supervised. Self-supervised variants incorporate constraints to enforce smoothness between neighboring samples in space, such as RegNeRF \cite{niemeyer2021Regnerf} (see Section~\ref{sec:regnerf}). InfoNeRF \cite{kim2021infonerf} prevents inconsistencies due to insufficient viewpoints by minimizing a ray entropy model and the \mbox{KL-divergence} between the normalized ray density obtained from neighbor viewpoints. In contrast, externally supervised regularization methods usually penalize differences with respect to extrinsic geometric cues. Depth-supervised NeRF~\cite{deng2021depth} encourages the rendered depth \eqref{eq:depth_nerf} to be consistent with a sparse set of 3D surface points obtained by structure from motion. A similar strategy is used in \cite{mari2022sat}, based on a set of 3D points refined by bundle adjustment; or \cite{roessle2021dense}, where a sparse point cloud is converted into dense depth priors by means of a depth completion network.

\section{Differential geometry to regularize NeRFs}
\label{sec:regnerf}

One of the major challenges when training a NeRF with insufficient data is to learn a consistent scene geometry so that the model extrapolates well on unseen views. In that case, it is common to add additional priors to the model to improve the quality of the learned models.

A classic hypothesis in depth and disparity estimation is that the target is smooth~\cite{scharstein2002taxonomy,hirschmuller2007stereo}. The same \textit{a priori} can be applied to the scene modeled by the NeRF.
Due to the ability of NeRFs to model transparent surfaces and volumes, the predicted weights can be highly irregular. As a consequence, it is easier to regularize across different rendered viewpoints (i.e. after projection onto a given camera) rather than directly regularizing the 3D scene itself. This means that instead of using the depth function $d$ from Eq.~\eqref{eq:depth_nerf}, it is more appropriate to work with the 
depth map $\tilde{d}$ produced by the NeRF model from a given viewpoint. This depth map $\tilde{d}$ is then indexed by its 2D coordinate $(x,y)$ instead of a ray in the 3D space.

In image processing, a classic way of enforcing smoothness is to add a regularization term in the loss function based on the gradients of the image.
For example, penalizing the squared $L_2$ norm of the gradients has the effect of removing high gradients in the depth map thus enforcing it to be smooth, as desired. In addition, it does not penalize slanted surfaces (since they have null Laplacian) as it would happen in the case of using a total variation regularization~\cite{rudin1992nonlinear}. 
The proposed regularization term thus reads
\begin{equation}
L_{depth} = \sum_{(x,y)} \text{clip}(\|\nabla \tilde{d}(x,y)\|^2, g_{max}).
\label{eq:tv_loss}
\end{equation}
In practice, we add a differentiable clipping to $L_{depth}$, parametrized by $g_{max}$, to preserve sharp edges that could otherwise be over-smoothed.

By \emph{only} changing the ReLU activation function to a Softplus activation function, the MLP used in NeRF can be transformed into a continuous and infinitely differentiable function similarly to \cite{gropp2020implicit}. This allows to train directly using the gradient of the model, or even higher order operators as shown later.

Traditionally, NeRFs are defined in terms of rays, which are {characterized} by an origin and a viewing direction $(\textbf{o}, \textbf{v})$. Consequently $d$ from~\eqref{eq:depth_nerf} is parameterized by $(\textbf{o}, \textbf{v})$ instead of the image coordinates $(x,y)$ as $\tilde{d}$ in~\eqref{eq:tv_loss}.
Let \mbox{$C: \mathbb{R}^2 \to \mathbb{R}^3$} be the transformation that converts the image coordinates into the equivalent ray so that $\tilde{d}(x,y) = d(\textbf{o},C(x,y))$ . Then the corresponding gradients are 
\begin{equation}
    \nabla_{(x,y)} \tilde{d}(x,y) = \textbf{J}_{C} (x,y) \nabla_{\textbf{v}} d(\textbf{o},\textbf{v}),
\end{equation}
where $\textbf{v} = C(x,y)$ and $\textbf{J}_{C}$ is the Jacobian matrix of $C$. This way, Eq.~\eqref{eq:tv_loss} can be expressed in terms of rays with the exception of $\textbf{J}_{C}$ that could be computed at the same time as the corresponding rays during the dataloading process. In practice, we use a simplified regularization loss that avoids computing $\textbf{J}_{C}$ (see Eq.~\eqref{eq:diffreg_loss}).

\myparagraph{Link with RegNeRF.}
In order to improve the robustness of NeRFs when training with few data, Niemeyer~\etal~\cite{niemeyer2021Regnerf} proposed RegNeRF, which also uses an additional term in the loss function to regularize the predicted depth map. This work additionally proposed an appearance regularization term using a normalizing flow network trained to estimate the likelihood of a predicted patch compared to normal patches from the JFT-300M dataset~\cite{sun2017revisiting}. While the later is not studied here, we show that their depth regularization term is simply an approximation of the more generic differential loss presented in Eq.~\eqref{eq:tv_loss}.

Consider the depth map $\tilde{d}$ and the set of coordinates $(x,y)$ that corresponds to the pixels of the depth map. RegNeRF regularizes depth by encouraging that neighboring pixels $(x+i,y+j)$ for $i,j \in \{0,1\}^2$ and $i+j=1$ have the same depth as the pixel $(x,y)$ such as
\begin{equation}
L_{depth} = \sum_{(x,y)} \sum_{\substack{(i,j) \in \{0,1\}^2 \\ i+j=1}} (\tilde{d}(x+i,y+j) - \tilde{d}(x,y))^2,
\label{eq:depth_regu_loss}
\end{equation}
which is a finite difference expression of the gradient of $\tilde{d}$. Thus the major difference between \eqref{eq:depth_regu_loss} and our approach is that \eqref{eq:depth_regu_loss} approximates the gradient with finite differences while we take advantage of automatic differentiation.

In practice, RegNeRF regularization is not done on the entire depth maps but rather by sampling patches.
The loss \eqref{eq:depth_regu_loss} is computed not only for all patches corresponding to a view in the training dataset, but also for rendered patches whose observation is not available. Indeed, all views should verify this depth regularity property, not only those in the training data. As a result,  RegNeRF requires modifying the dataloaders to incorporate patch-based sampling and rays corresponding to unseen views.  
Note that our depth regularization term \eqref{eq:tv_loss} does not require patches and therefore can be directly applied using single rays sampling as traditionally done to train NeRFs. It also does not regularize unseen views as explained in Section~\ref{sec:exps}.

\myparagraph{Normals regularization.}
The regularization term \eqref{eq:tv_loss} relies on depth maps. However, differential geometry also allows us to regularize other geometry-related features when training a NeRF. For example, consider $n$, the function that returns the scene normals for a given ray, whose projection, or map of normals, is denoted $\tilde{n}$. In that case, the regularization of the normals of the scene becomes
\begin{equation}
L_{normals} = \sum_{(x,y)} \|J_{\tilde{n}}(x,y)\|_F^2
\label{eq:diffreg_normal_loss}
\end{equation}
where $J_{\tilde{n}}$ is the Jacobian of the map of normals. This regularizer was applied to generate one of the results in Fig.~\ref{fig:teaser}.

\begin{figure}
    \centering
    \includegraphics[width=\linewidth]{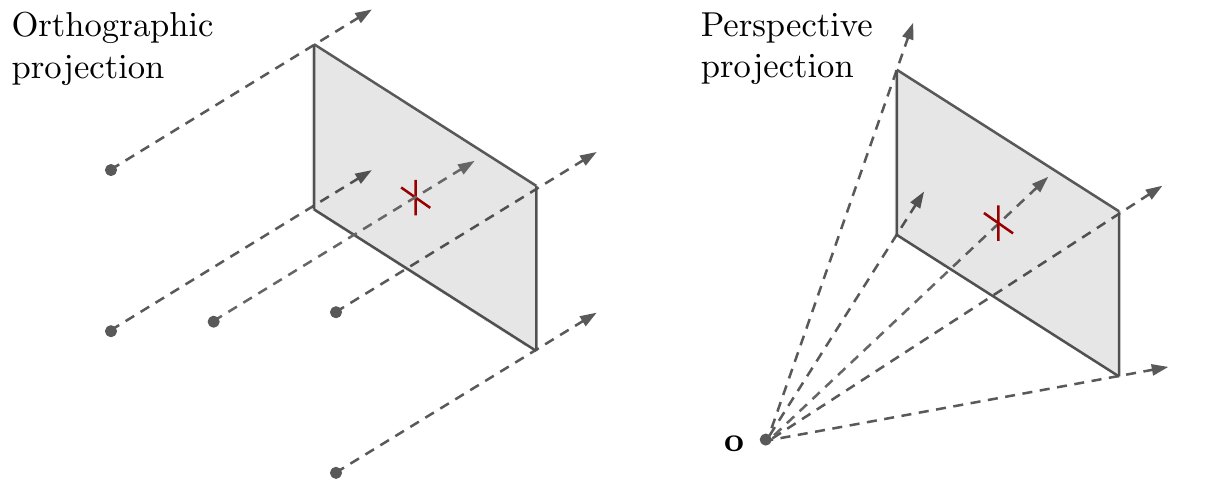}
\caption{All perspective projection rays originate at the same center of projection \textbf{o}, located at a finite distance from the image plane. The center of projection in orthographic projection is at infinity, which can be represented by using a different origin for each ray, so that the origin points are parallel to the image plane.}
    \label{fig:cam_models}
\end{figure}

\myparagraph{Simplified regularization loss.}
The main problem with the loss presented in Eq.~\eqref{eq:tv_loss} is that it does not depend only on each individual ray, but also requires additional camera information to compute $\textbf{J}_C$. Since this can be unpractical depending on the camera model, we propose to use a different and fixed local camera model only for the regularization process. Instead of using the usual perspective projection models associated with the training data, it is possible to regularize the scene as if the ray being processed originated from an orthographic projection camera, as illustrated in Fig.~\ref{fig:cam_models}.

Consider a ray defined by its origin $\textbf{o}$ and its direction $\textbf{v})$.
Let $(\textbf{i},\textbf{j})$ be a local orthonormal basis of the plane defined by $\textbf{o}$ and $\textbf{v}$. Using an orthographic projection camera, the direction is fixed and only the origin changes to obtain other rays from the same camera. Therefore $C$, defined such that $\tilde{d}(x,y) = d(C(x,y), \textbf{v})$, is explicit and $C(x,y) = x\textbf{i} + y\textbf{j}$. This leads to 
\begin{equation}
    \textbf{J}_{C} (x,y) = \begin{pmatrix} \textbf{i} \\ \textbf{j} \end{pmatrix} \in \mathbb{R}^{2 \times 3}.
\end{equation}
Therefore 
\begin{equation}
\nabla_{(x,y)} \tilde{d}(x,y) = \begin{pmatrix} \langle\nabla_{\textbf{o}} d(\textbf{o},\textbf{v}), \textbf{i}\rangle \\ \langle\nabla_{\textbf{o}} d(\textbf{o},\textbf{v}), \textbf{j}\rangle \end{pmatrix}
\end{equation}
and $\|\nabla \tilde{d}(x,y)\|^2 = \langle\nabla_{\textbf{o}} d(\textbf{o},\textbf{v}), \textbf{i}\rangle^2 + \langle\nabla_{\textbf{o}} d(\textbf{o},\textbf{v}), \textbf{j}\rangle^2$. Since $(\textbf{i}, \textbf{j}, \textbf{v})$ is, by construction, an orthonormal basis of the space, we also have that $\|\nabla_{\textbf{o}} d(\textbf{o},\textbf{v})\|^2 = \langle\nabla_{\textbf{o}} d(\textbf{o},\textbf{v}), \textbf{i}\rangle^2 + \langle\nabla_{\textbf{o}} d(\textbf{o},\textbf{v}), \textbf{j}\rangle^2 + \langle\nabla_{\textbf{o}} d(\textbf{o},\textbf{v}), \textbf{v}\rangle^2$ thus 
\begin{align}
L_{depth} &= \sum_{(\textbf{o}, \textbf{v}) \in \mathcal{R}} \|\nabla_{\textbf{o}} d(\textbf{o}, \textbf{v}) \|^2 - \langle\nabla_{\textbf{o}} d(\textbf{o}, \textbf{v}),  \textbf{v}\rangle^2\\
&= \sum_{(\textbf{o}, \textbf{v}) \in \mathcal{R}} \|\nabla_{\textbf{o}} d(\textbf{o}, \textbf{v})  - \langle\nabla_{\textbf{o}} d(\textbf{o}, \textbf{v}),  \textbf{v}\rangle\textbf{v}\|^2.
\label{eq:diffreg_loss}
\end{align}
Note how Eq.~\eqref{eq:diffreg_loss} does not depend on the choice of $(\textbf{i}, \textbf{j})$, is entirely defined by the knowledge of the ray $(\textbf{o}, \textbf{v})$ and is independent from $\textbf{J}_{C}(x,y)$.

\section{Experimental results}
\label{sec:exps}

\begin{table*}
    \centering
  {\small
  \begin{tabular}{lccccccccccc}
    \toprule
    &               &   fern   &  flower & fortress &   horns &  leaves &  orchids &   room  &   trex  & avg. \\
    \midrule
    \multirow{6}{*}{\rotatebox[origin=c]{90}{PSNR}}
    & PixelNeRF ft~\cite{yu2021pixelnerf}  &     -   &     -   &     -   &     -   &     -   &     -   &     -   &     -   &    16.17\\
    & SRF ft~\cite{chibane2021stereo}      &     -   &     -   &     -   &     -   &     -   &     -   &     -   &     -   &    17.07\\
    & MVSNeRF ft~\cite{chen2021mvsnerf}    &     -   &     -   &     -   &     -   &     -   &     -   &     -   &     -   &    17.88\\
    & RegNeRF (w/o app. reg.) &   19.85 &   19.64 &   22.28 &   13.05 &   16.46 &   15.43  &   20.62 &   20.37 &   18.46 \\
    & DiffNeRF (ours)                         &\B{20.15}&\B{20.27}&\B{23.68}&\B{17.80}&\B{16.88}&   15.50  &   21.04 &\B{20.58}&\B{19.49}\\
    & RegNeRF~\cite{niemeyer2021Regnerf}   &   19.87 &   19.93 &   23.32 &   15.65 &   16.60 &\B{15.56} &\B{21.53}&   20.17 &   19.08 \\
    \midrule
    \multirow{3}{*}{\rotatebox[origin=c]{90}{SSIM}}
    & RegNeRF (w/o app. reg.) &   0.694 &   0.677 &   0.706 &   0.486 &   0.599 &   0.483 &   0.843 &   0.774 &   0.658 \\
    & DiffNeRF (ours)                         &\B{0.703}&\B{0.707}&\B{0.761}&\B{0.680}&\B{0.645}&   0.487 &\B{0.864}&\B{0.791}&\B{0.705}\\
    & RegNeRF~\cite{niemeyer2021Regnerf}   &   0.697 &   0.688 &   0.743 &   0.610 &   0.613 &\B{0.502}&   0.861 &   0.766 &   0.685 \\
    \midrule
    \multirow{3}{*}{\rotatebox[origin=c]{90}{LPIPS}}
    & RegNeRF (w/o app. reg.) &   0.323 &   0.243 &   0.294 &   0.341 &   0.229 &   0.259 &   0.204 &   0.197 &   0.261 \\
    & DiffNeRF (ours)                         &\B{0.290}&\B{0.223}&\B{0.219}&\B{0.293}&\B{0.186}&\B{0.247}&\B{0.171}&\B{0.166}&\B{0.224}\\
    & RegNeRF~\cite{niemeyer2021Regnerf}   &   0.304 &   0.234 &   0.258 &   0.356 &   0.222 &   0.251 &   0.185 &   0.197 &   0.251 \\
    \bottomrule
  \end{tabular}
  }
\vspace{-.5em}
  \caption{Quantitative comparison of novel view synthesis for different NeRF regularization on the LLFF dataset. All models were trained using \ul{only three input views}. \emph{RegNeRF (w/o app. reg.)} corresponds to the original RegNeRF without appearance regularization, while the proposed framework is \emph{DiffNeRF}. The results using RegNerf with appearance regularization are also provided for reference. The proposed regularization almost systematically achieves the best results across all metrics without requiring any additional appearance regularization. The LPIPS metric is computed using the official implementation provided by Zhang~\etal~\cite{zhang2018perceptual}. Best results are shown in bold.}
  \label{tab:regnerf_llff}
\end{table*}

\begin{table*}
  \renewcommand{\arraystretch}{1.3}
  \centering
  \resizebox{\linewidth}{!}{
  \begin{tabular}{lcccccccccccccccc}
    \toprule
   &   8   &  21   &  30   &  31   &   34  &   38  &   40  &   41  &   45  &   55  &   63  &   82  &  103  &  110  &  114  &  avg  \\
   \midrule
   \makecell{RegNeRF\\ (w/o app. reg.)} &   19.06 &   12.42 &   22.45 &   16.35 &   18.13 &   16.92 &   18.63 &   15.97 &   16.29 &   17.75 &   20.57 &   17.54 & 22.10 &   17.97 &   21.31 &   18.23 \\
   DiffNeRF (ours)                      &   15.47 &\B{13.63}&\B{23.18}&   16.74 &\B{18.66}&\B{17.28}&   18.57 &   15.53 &\B{16.45}&\B{17.94}&   21.65 &   15.19 &\B{23.69}&\B{20.32}& 21.41 &   18.38 \\
   RegNeRF~\cite{niemeyer2021Regnerf}   &\B{19.45}&   12.76 &   22.92 &\B{16.84}&   18.24 &   17.12 &\B{19.09}&\B{18.41}&   16.44 &   17.61 &\B{22.91}&\B{19.42}& 22.95 &   18.06 &\B{21.52}&\B{18.92}\\
   \bottomrule
  \end{tabular}
  }
 \vspace{-.5em}
  \caption{Quantitative comparison of novel view synthesis for different NeRF regularization on the DTU dataset. All models were trained using \ul{only three input views}. \emph{RegNeRF (w/o app. reg.)} corresponds to the original RegNeRF without appearance regularization, while the proposed framework is \emph{DiffNeRF}. The results using RegNerf with appearance regularization are also provided for reference. The case of scenes 41 and 82 are discussed in Section~\ref{sec:exps}. Best results are shown in bold. }
  \label{tab:regnerf_dtu}
\end{table*}

We test the impact of the proposed differential regularization on the task of scene estimation using only three input views. This is the most extreme test case and, as such, it is highly reliant on the quality of the regularization to avoid catastrophic collapse as shown by Niemeyer~\etal~\cite{niemeyer2021Regnerf} for mip-NeRF~\cite{barron2021mipnerf}.  
In order to compare the proposed formalization of RegNeRF~\cite{niemeyer2021Regnerf} to its original version, we modified the code of the authors and replaced their depth loss by the one in~\eqref{eq:diffreg_loss}. We refer to our approach as DiffNeRF. The code to reproduce the results is available at \url{https://github.com/tehret/diffnerf}.

\myparagraph{Results on LLFF~\cite{mildenhall2019local}.}
In Table~\ref{tab:regnerf_llff}, we compare the results of the original RegNeRF (using the models trained by the authors) with our DiffNeRF formalization~\eqref{eq:diffreg_loss}. Since the code released by the authors does not contain the additional appearance loss, we added another comparison that corresponds to RegNeRF without the additional appearance regularization (\ie training from scratch using the available code). The proposed DiffNeRF not only improves by 1dB the PSNR of reconstructed unseen views compared to the equivalent RegNeRF version, it also outperforms RegNeRF with appearance regularization by almost 0.5dB. This is also the case for other metrics such as SSIM and LPIPS.

\begin{figure*}
    \centering
    \resizebox{\linewidth}{!}{
    \small
    \begin{tabular}{c@{\hskip 0.01\linewidth}c@{\hskip 0.01\linewidth}c@{\hskip 0.01\linewidth}c@{\hskip 0.01\linewidth}c@{}}
        \vspace{0.5em}
        & RegNeRF~\cite{niemeyer2021Regnerf} &RegNeRF (w/o app. reg.) &  DiffNeRF (ours) & Ground truth \\
    \rotatebox[origin=c, x=-1cm]{90}{horns} & \includegraphics[width=0.24\linewidth]{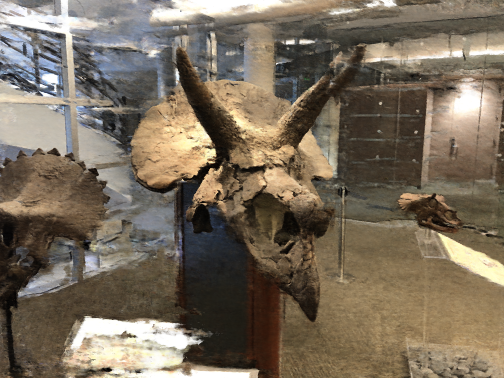} &
    \includegraphics[width=0.24\linewidth]{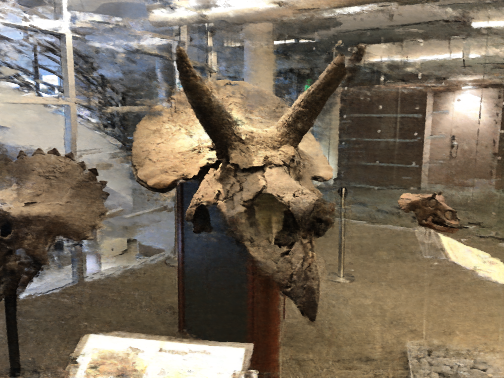} &
    \includegraphics[width=0.24\linewidth]{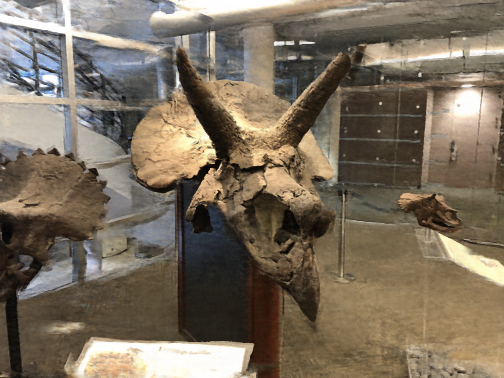} &
    \includegraphics[width=0.24\linewidth]{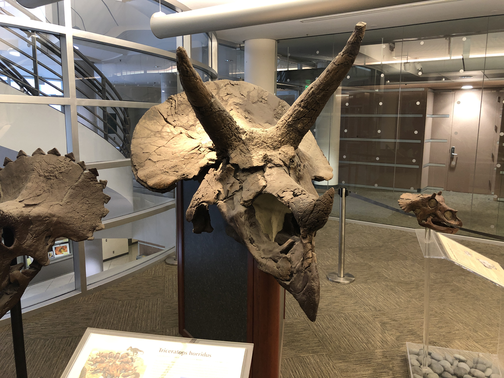} \\
    & \includegraphics[width=0.24\linewidth]{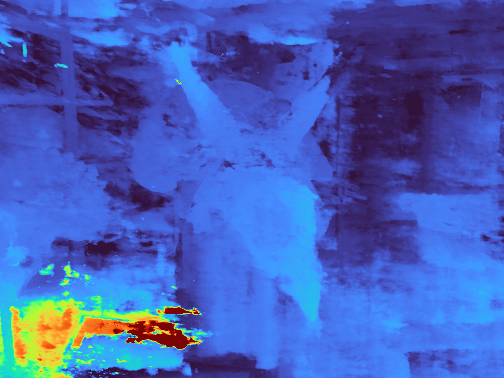} &
    \includegraphics[width=0.24\linewidth]{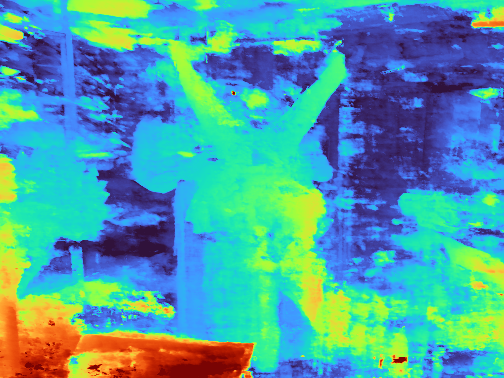} &
    \includegraphics[width=0.24\linewidth]{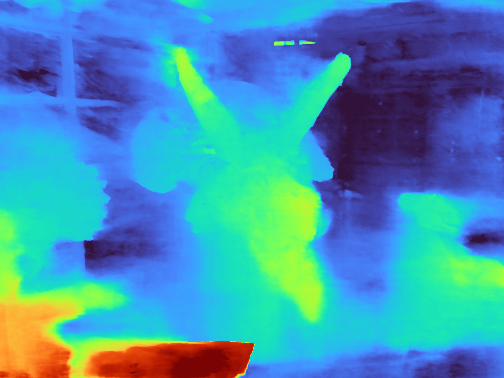} &
    \includegraphics[width=0.24\linewidth]{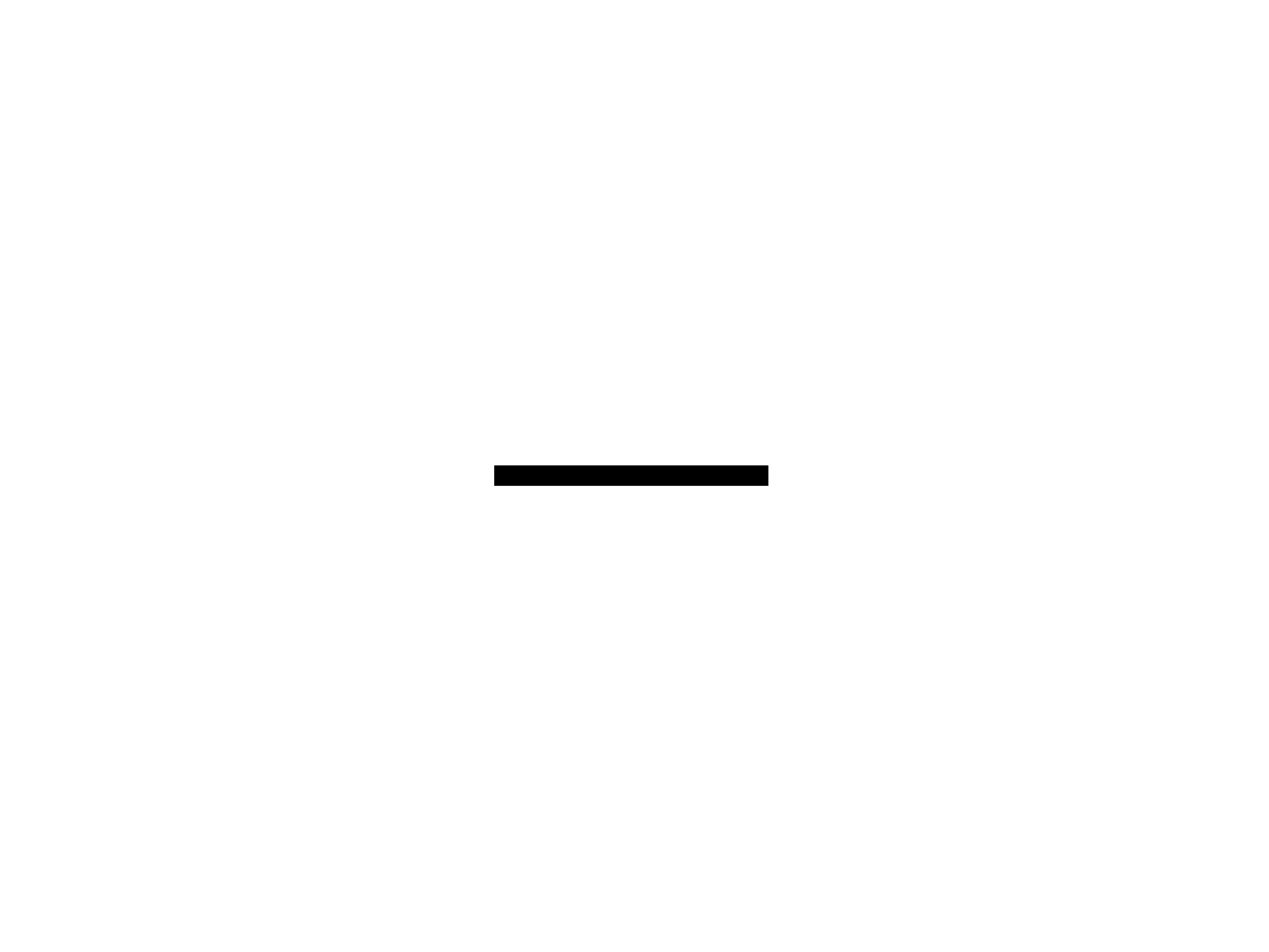} \\
    \rotatebox[origin=c, x=-1.2cm]{90}{trex} & \includegraphics[width=0.24\linewidth]{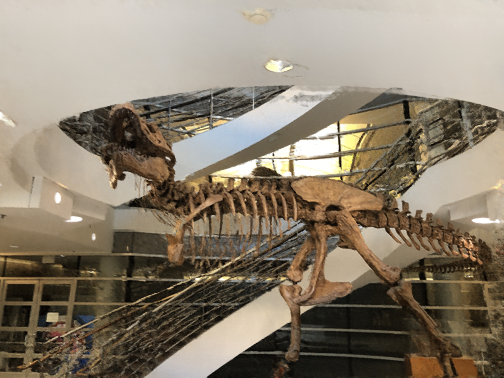} &
    \includegraphics[width=0.24\linewidth]{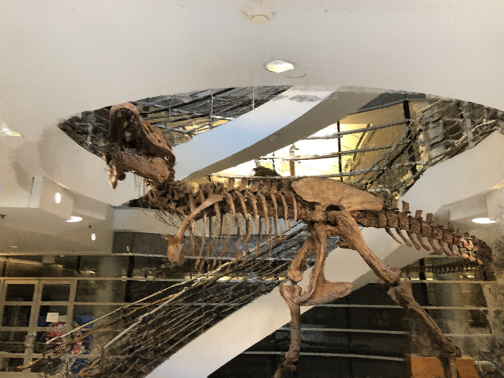} &
    \includegraphics[width=0.24\linewidth]{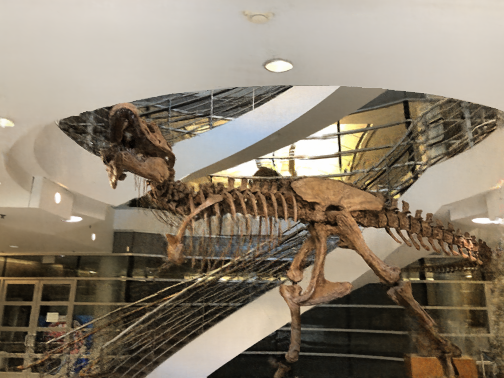} &
    \includegraphics[width=0.24\linewidth]{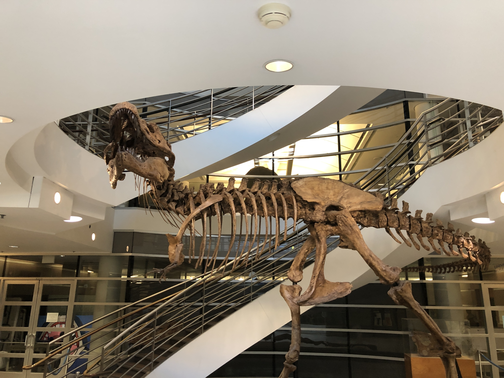} \\
    & \includegraphics[width=0.24\linewidth]{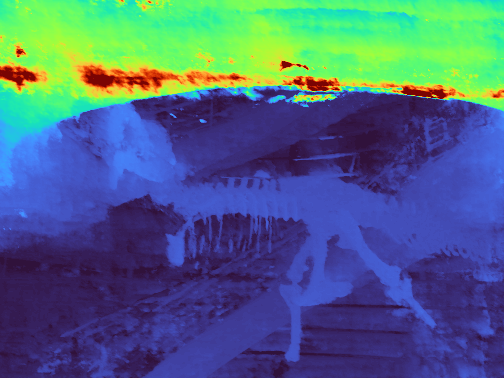} &
    \includegraphics[width=0.24\linewidth]{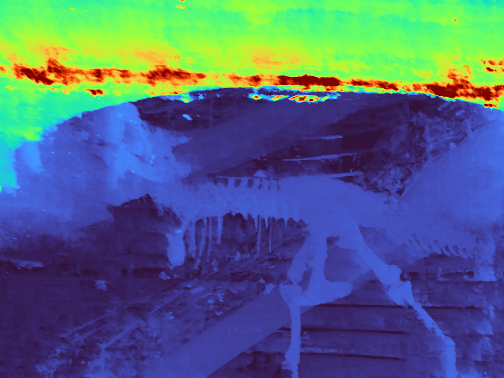} &
    \includegraphics[width=0.24\linewidth]{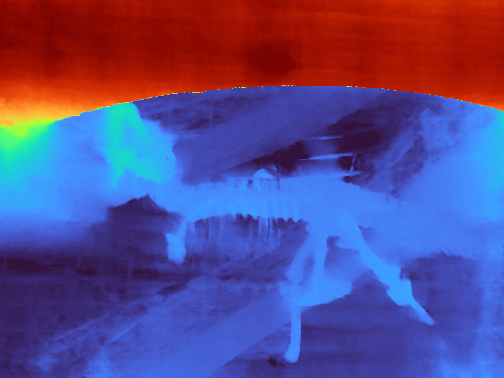} &
    \includegraphics[width=0.24\linewidth]{images/llff/blank_llff.pdf} \\
    \end{tabular}
    }
    \vspace{-1em}
    \caption{Visual examples of novel view synthesis for the \emph{horns} (top) and \emph{trex} (bottom) sequences of the LLFF dataset after training with \ul{three views}. The depth map produced by the proposed DiffNeRF is more regular than those produced RegNeRF. It also recovers more details both in the foreground (see the sign panel on the left or the triceratops' left horn) but also in the background (see the glass panels and the handrails).}
    \label{fig:regnerf_llff}
\end{figure*}

Visual results on two examples of the LLFF dataset are shown in Fig.~\ref{fig:regnerf_llff}. In both cases, we compare the proposed version with the models trained by Niemeyer~\etal~\cite{niemeyer2021Regnerf}. The \textit{horns} scene in Fig.~\ref{fig:regnerf_llff} shows a first example where our formalization outperforms RegNeRF across all evaluation metrics. The proposed method is able to learn a better geometry of the image, leading to a more complete reconstruction of the triceratops skull (see the horn on the right), but also of the rest of the scene, such as the sign panel in the foreground or the handrails in the background. Similar improvements can be observed in the \textit{trex} scene.

Fig.~\ref{fig:teaser} shows another result, with the \textit{room} scene of the LLFF dataset where the PSNR obtained with DiffNeRF is worse with respect to RegNeRF with appearance regularization. However, the depth map estimated by our formalism is still much smoother without losing details.  In addition, as in the triceratops example, we can see that some details are also better reconstructed, like the audio conferencing system in the middle of the table. The LPIPS metric in Table~\ref{tab:regnerf_llff} also seems to confirm that the DiffNeRF results present less visual artifacts than RegNeRF.
Both Fig.~\ref{fig:teaser} and Fig.~\ref{fig:regnerf_llff} show that the DiffNeRF depth maps are better regularized than the original RegNeRF. In DiffNeRF we only use the input views at training time, 
without regularizing unseen views 
or requiring patch-based dataloaders with a predefined patch size (as in RegNeRF). This shows that the proposed formalism yields a better generalization.
All experiments with LLFF were computed using a weight of $2e^{-4}$ for the regularization term with a clipping value $g_{max}=20$. %

\begin{figure*}
    \centering
    \resizebox{\linewidth}{!}{
    \small
    \begin{tabular}{c@{\hskip 0.01\linewidth}c@{\hskip 0.01\linewidth}c@{\hskip 0.01\linewidth}c@{\hskip 0.01\linewidth}c@{}}
        \vspace{0.5em}
        &RegNeRF~\cite{niemeyer2021Regnerf} & RegNeRF (w/o app. reg.) & DiffNeRF (ours) & Ground truth \\
    \rotatebox[origin=c, x=-1cm]{90}{scan30} &
    \includegraphics[width=0.24\linewidth]{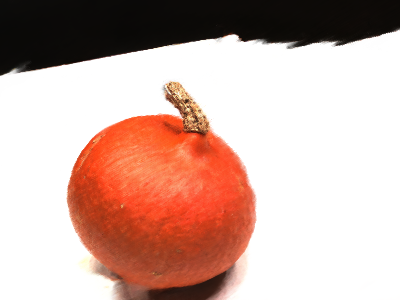} &
    \includegraphics[width=0.24\linewidth]{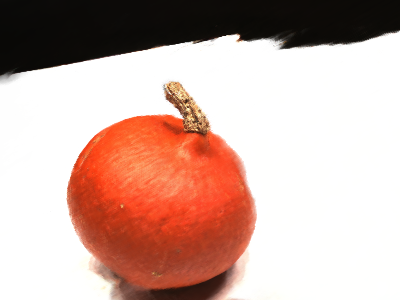} &
    \includegraphics[width=0.24\linewidth]{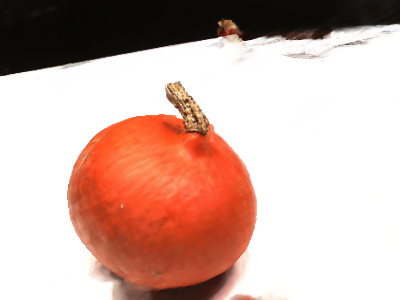} &
    \includegraphics[width=0.24\linewidth]{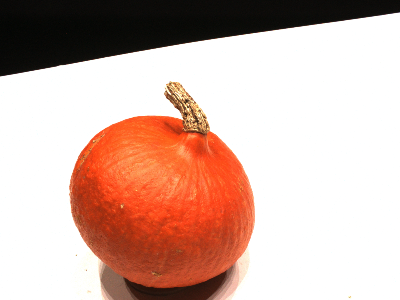} \\
    &\includegraphics[width=0.24\linewidth]{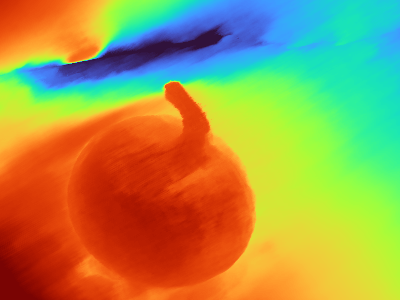} &
    \includegraphics[width=0.24\linewidth]{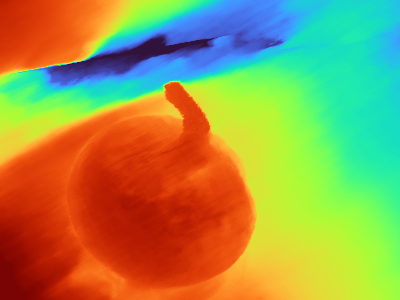} &
    \includegraphics[width=0.24\linewidth]{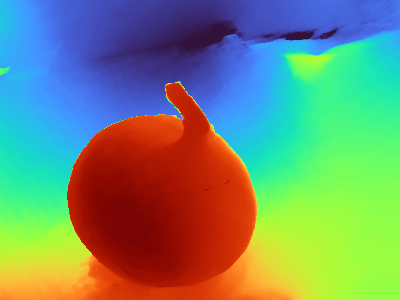} &
    \includegraphics[width=0.24\linewidth]{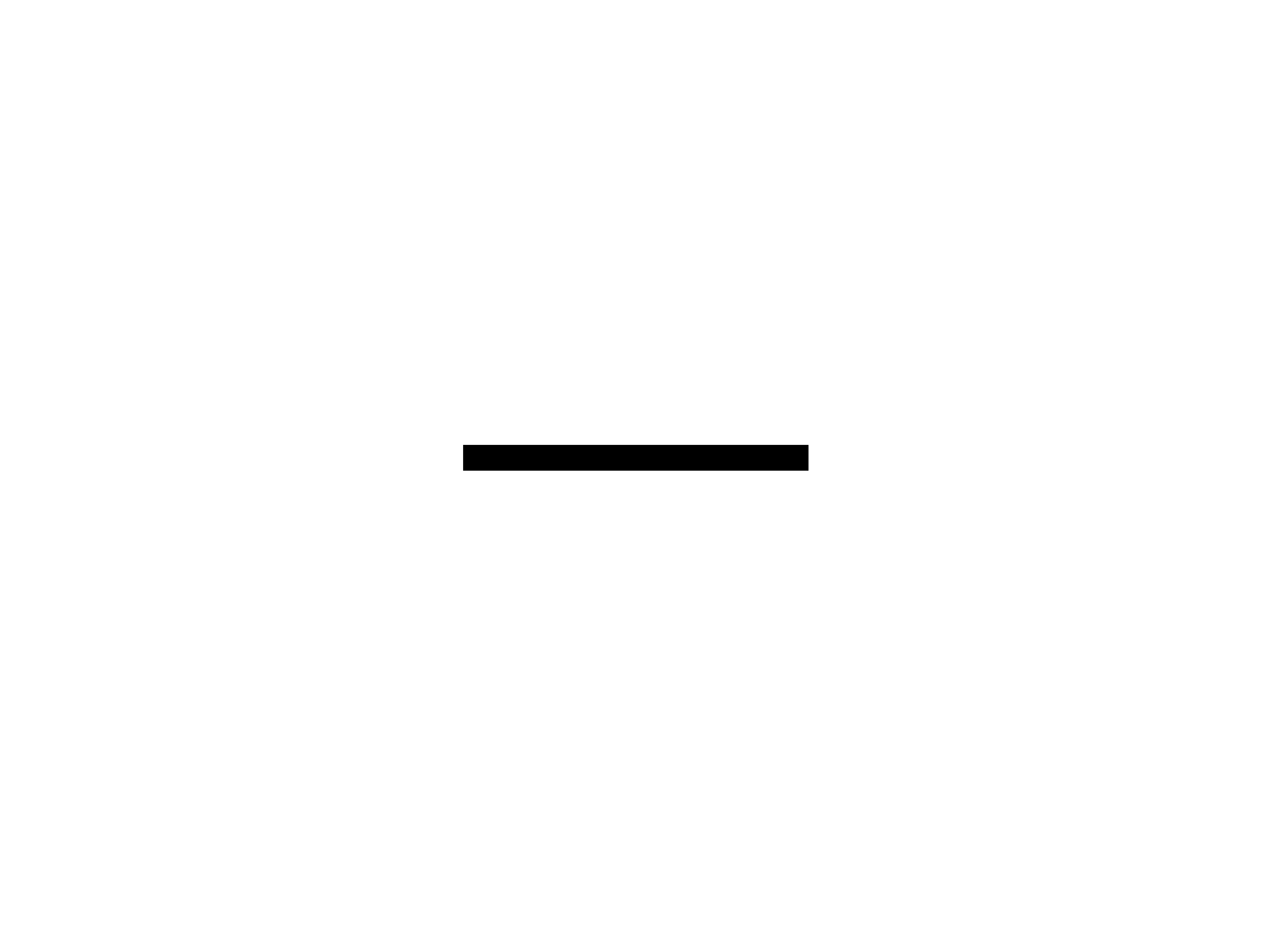} \\
    \rotatebox[origin=c, x=-1cm]{90}{scan40} &
    \includegraphics[width=0.24\linewidth]{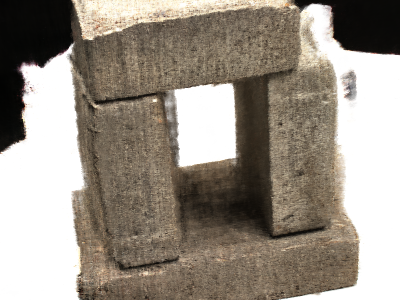} &
    \includegraphics[width=0.24\linewidth]{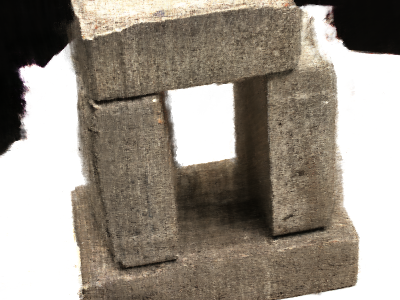} &
    \includegraphics[width=0.24\linewidth]{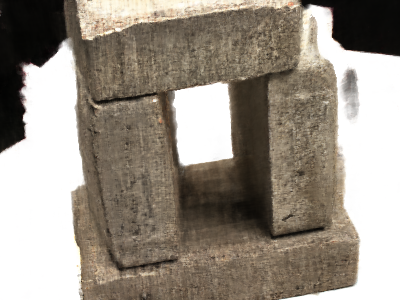} &
    \includegraphics[width=0.24\linewidth]{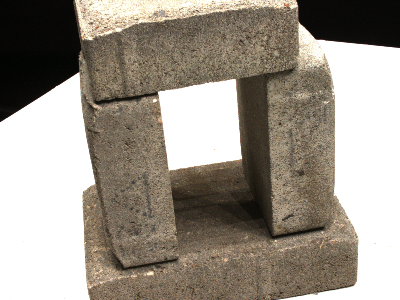} \\
    &\includegraphics[width=0.24\linewidth]{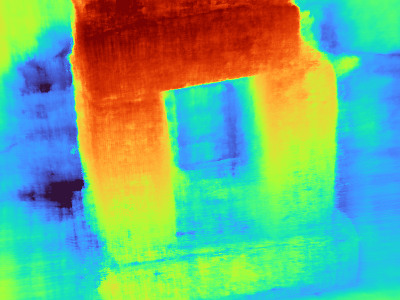} &
    \includegraphics[width=0.24\linewidth]{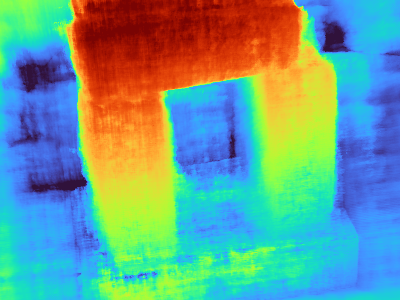} &
    \includegraphics[width=0.24\linewidth]{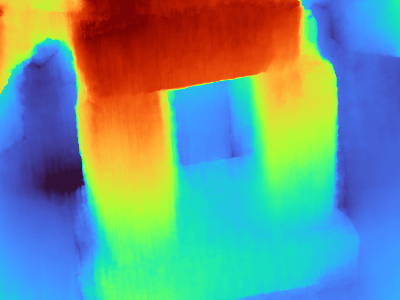} &
    \includegraphics[width=0.24\linewidth]{images/dtu/blank.pdf} \\
    \end{tabular}
    }
    \vspace{-1em}
    \caption{Visual example of novel view synthesis for scenes 30 and 40 of the DTU dataset after training with \ul{three views}. The depth map produced by the proposed DiffNeRF is more regular than those produced RegNeRF. It also separates better the object from the background.}
    \label{fig:regnerf_dtu}
\end{figure*}

\myparagraph{Results on DTU~\cite{jensen2014large}.}
Table~\ref{tab:regnerf_dtu} and Fig.~\ref{fig:regnerf_dtu} present results on the DTU dataset. Again, DiffNeRF produces results with a smoother scene geometry. All experiments with DTU were computed using a weight of $2e^{-4}$ with a clipping value $g_{max}=5$.

\myparagraph{Parameters study.} We illustrate in Fig.~\ref{fig:params_study} the impact of the two parameters of the proposed regularization: the weight of the regularization term in the loss and the value of the clipping. When the regularization is too weak, the surface exhibits irregular patterns. On the contrary, a regularization that is too strong can make details disappear (for example when parts of the pumpkin are merged with the background). A strong clipping allows to get back some details but can lead to visual artifacts such as staircasing. The proposed set of parameters leads to a smooth surface while keeping details.

\tikzset{
     fig_label/.style={
        draw=black,
        fill=white,
        text=black
    }
}

\begin{figure}
    \centering
    \begin{tabular}{@{}c@{\hskip 0.01\linewidth}c@{}}
    \begin{tikzpicture}
    \node[anchor=south west, inner sep=0] (img) at (0,0){
    \includegraphics[width=0.49\linewidth]{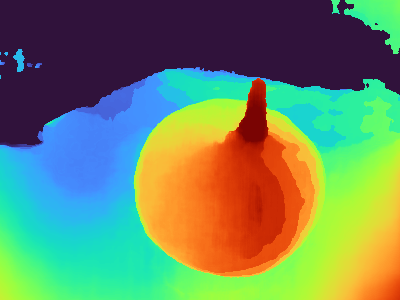}};
    \begin{scope}[x={(img.south east)},y={(img.north west)}]
     \node[fig_label, text width=0.4cm] at (0.1,0.88) {(a)};
    \end{scope}
    \end{tikzpicture} &
    \begin{tikzpicture}
    \node[anchor=south west, inner sep=0] (img) at (0,0){
    \includegraphics[width=0.49\linewidth]{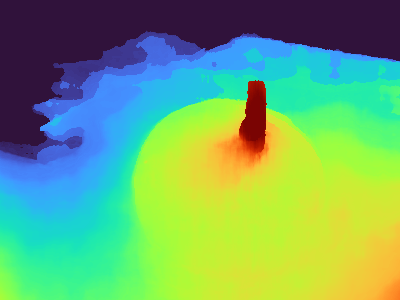}};
    \begin{scope}[x={(img.south east)},y={(img.north west)}]
     \node[fig_label, text width=0.4cm] at (0.1,0.88) {(b)};
    \end{scope}
    \end{tikzpicture}
    \\
    \begin{tikzpicture}
    \node[anchor=south west, inner sep=0] (img) at (0,0){
    \includegraphics[width=0.49\linewidth]{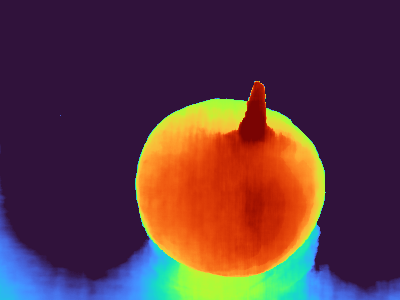}};
    \begin{scope}[x={(img.south east)},y={(img.north west)}]
     \node[fig_label, text width=0.4cm] at (0.1,0.88) {(c)};
    \end{scope}
    \end{tikzpicture}
    &
    \begin{tikzpicture}
    \node[anchor=south west, inner sep=0] (img) at (0,0){
    \includegraphics[width=0.49\linewidth]{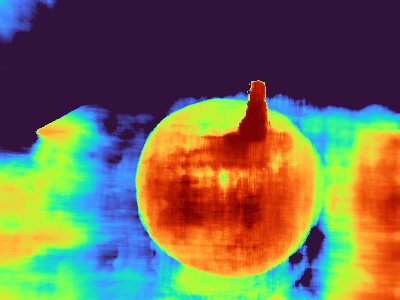}};
     \begin{scope}[x={(img.south east)},y={(img.north west)}]
     \node[fig_label, text width=0.4cm] at (0.1,0.88) {(d)};
    \end{scope}
    \end{tikzpicture}\\
    \end{tabular}
        \vspace{-.5em}
    \caption{Visual impact of the two parameters of the regularization (reconstructions from three views). (a) strong regularization and clipping, (b) strong regularization and little clipping, (c) medium regularization and clipping, (d) little regularization and clipping.}
    \label{fig:params_study}
\end{figure}

\begin{figure}
    \centering
    \begin{tabular}{@{}c@{\hskip 0.01\linewidth}c@{}}
    \includegraphics[width=0.49\linewidth]{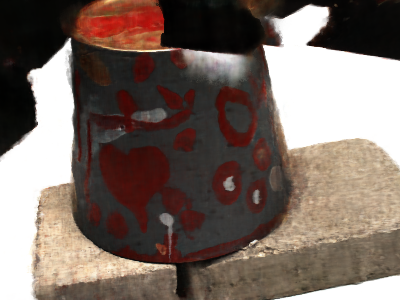} &
    \includegraphics[width=0.49\linewidth]{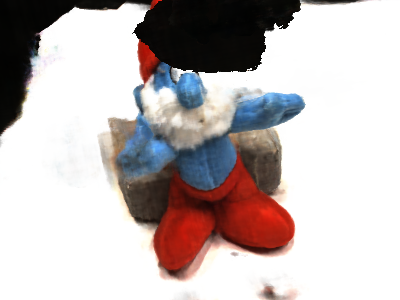} \\
    \includegraphics[width=0.49\linewidth]{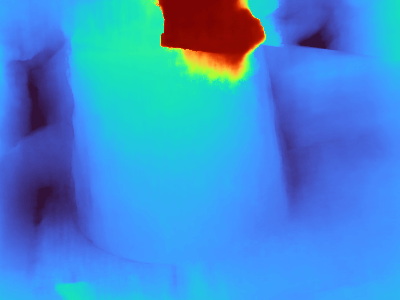} &
    \includegraphics[width=0.49\linewidth]{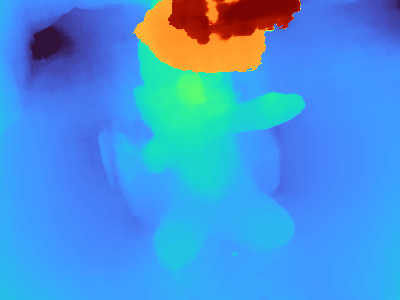} \\
    \end{tabular}
    \vspace{-.5em}
    \caption{Failure cases for scenes 41 and 82 of the DTU dataset reconstructed from three views. "Floaters" (groups of points with a non zero density and disjoint from the scene) hide portions of the scene when synthesizing novel views.}
    \label{fig:failure_cases}
\end{figure}

\myparagraph{Limitations.}
During the experiments, we observed two main limitations. The first one is the apparition of "floaters", groups of points with a non zero density and disjoint from the scene. These floaters can hide portions of the scene when synthesizing novel views (see Fig.~\ref{fig:failure_cases}). In the DTU dataset, we find these floating artifacts to be related to the large textureless background regions or areas observed by a single camera (\ie when the problem is not well defined, note that these regions are still regularized). We did not observe such floaters in the LLFF dataset. This also explains why regularizing unseen views, \ie without any data attachment term, is not a good idea since it encourages the creation of such floaters.

The second limitations is the computation performance. Since the proposed regularization requires computing gradients, it is expected to be slower and require more memory. Nevertheless, since there is no need to regularize unseen views, the proposed method remains competitive (for the depth regularization). A comparison is shown in Table~\ref{tab:comp_time}.

\begin{table}
\centering
  \begin{tabular}{lcc}
    \toprule
   &   Rays/s & Batch size\\
   \midrule
   RegNeRF~\cite{niemeyer2021Regnerf} &   $\sim 6000$ & $\sim 2000$\\
   DiffNeRF (depth reg.)              &   $\sim 5000$ & $\sim 1000$\\
   DiffNeRF (normals reg.)            &   $\sim 1100$ & $\sim 250$\\
   \midrule
  \end{tabular}
    \vspace{-1em}
  \caption{Computation speed (in rays per seconds) for the different methods on a NVIDIA V100 16Go GPU. Because diffNeRF does not require sampling additional rays from unseen views, the computation is barely slower than RegNeRF~\cite{niemeyer2021Regnerf} ($\sim 16\%$). Higher order regularization (such as normals regularization) are much slower though.}
  \label{tab:comp_time}
\end{table}

\section{Extension to surface regularization using mean and Gaussian curvatures}
\label{sec:surface_reg}

Another trend with NeRF-like models is to directly learn a surface model instead of a density function as shown in Section~\ref{sec:pres_nerf}. In particular, IDR~\cite{yariv2020multiview} and VolSDF~\cite{yariv2021volume} both learn the surface {by means of} a signed distance function (SDF). This SDF can then be used {in a direct manner} or as a guide to sample points as done in NeRF to learn the surface. Since this SDF  can be seen as an implicit function $F$ defining the surface $\mathcal{S}$ as the set of points $\left\{\textbf{x} \in \mathbb{R}^3~|~F(x)=0\right\}$, it is possible to compute other differential quantities related to surface regularity, such as the curvature. 
This allows to directly regularize the surface instead of regularizing the projections of the scene as shown in Section~\ref{sec:regnerf}.
We propose in this section to look at the Gaussian curvature $\gamma_{gauss}$ and the mean curvature $\gamma_{mean}$, since they both have an analytical expression that can be easily implemented using the existing deep learning frameworks.

These curvatures are respectively defined as
\begin{equation}
\gamma_{mean} = -\textbf{div}\left( \frac{\nabla F}{\|\nabla F\|} \right)
\label{eq:curv_1}
\end{equation}
and
\begin{equation}
\gamma_{gauss} = \frac{\nabla F \times H^*(F) \times \nabla F^t}{\|\nabla F\|^4}, 
\label{eq:curv_2}
\end{equation}
where $H^*$ is the adjoint of the Hessian of $F$. Derivation details of these two curvatures can be found in~\cite{goldman2005curvature}. Using \eqref{eq:curv_1} and \eqref{eq:curv_2}, we can define a regularization loss similar to the one presented in Section~\ref{sec:regnerf} as
\begin{equation}
L_{curv}(\kappa_{curv}) = \mathbb{E}_{x\in \mathcal{S}}\left[\min\left(|\gamma(x)|, \kappa_{curv}\right)\right],
\label{eq:curvs_loss}
\end{equation}
where $\gamma$ can be either $\gamma_{mean}$ or $\gamma_{gauss}$, depending on the preferred behavior, and $\kappa_{curv}$ is a clipping value. %
The final loss to train %
a regularized VolSDF model using~\eqref{eq:curvs_loss} becomes
\begin{equation}
L = L_{RGB} + \lambda_{SDF} L_{SDF} + \lambda_{curv} L_{curv}(\kappa_{curv}) 
\label{eq:curvs_reg_loss_volsdf}
\end{equation}
with
\begin{equation}
L_{SDF} = \mathbb{E}_{x \in \mathbb{R}^3} \left[(\|\nabla F(x)\| - 1 )^2\right].
\end{equation}
As in~\cite{yariv2021volume}, the $L_{SDF}$ term enforces the Eikonal constraint on the implicit function $F$, thus learning a signed distance function.
Note that~\eqref{eq:curvs_reg_loss_volsdf} makes it possible to regularize the surface directly during training instead of doing it in different stages as in \cite{yang2021geometry}.

\begin{figure}
    \centering
    \small
    \includegraphics[width=0.32\linewidth]{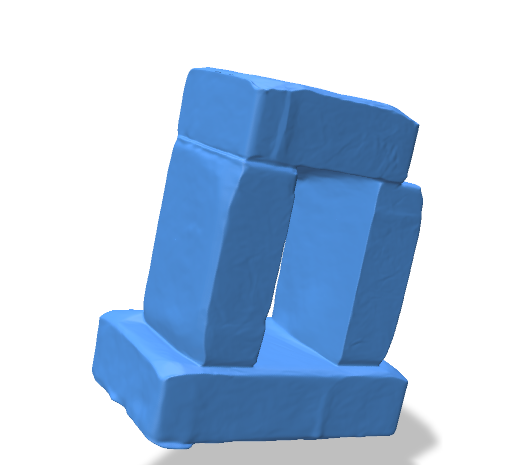}
    \includegraphics[width=0.32\linewidth]{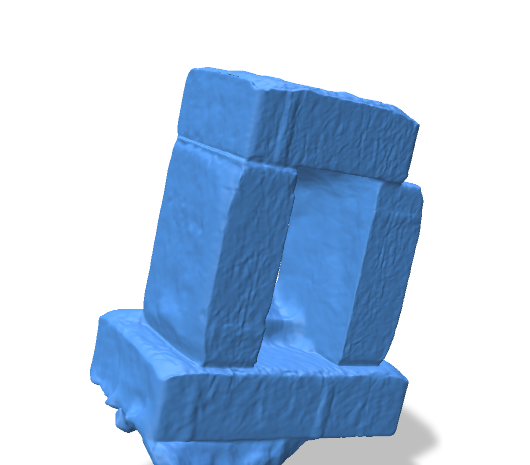}
    \includegraphics[width=0.31\linewidth]{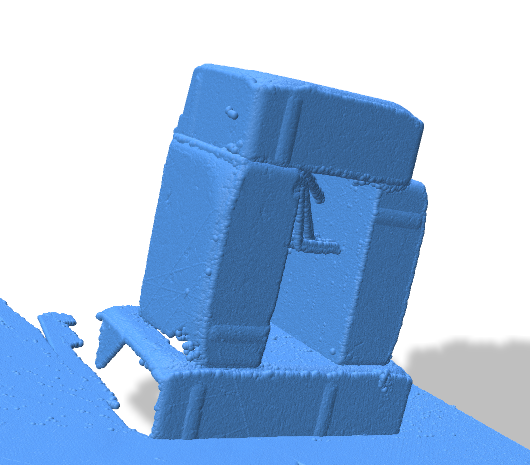}
    
    \caption{Visual example of a regularized reconstruction of the scene 40 of the DTU dataset. From left to right: regularized reconstruction using Gaussian curvature~\eqref{eq:curv_2}, original VolSDF results and ground truth.}
    \label{fig:exper_curvature}
\end{figure}

The regularization is characterized by the same two parameters, the regularization weight and the clipping value, than the regularization presented in Section~\ref{sec:regnerf}. To understand the impact of these parameters, we refer to the definition of the mean and Gaussian curvature in terms of the minimum curvature $\gamma_{min}$ and maximum curvature $\gamma_{max}$ of the surface at a given point
\begin{equation}
\gamma_{mean} = \frac{\gamma_{min} + \gamma_{max}}{2} \hspace{0.5em}\text{ and }\hspace{0.5em} \gamma_{gauss} = \gamma_{min} \gamma_{max}.
\label{eq:curv_def_min_max}
\end{equation}
Although this is not a practical definition of the curvature, since it does not allow for direct computation, it shows that minimizing the mean curvature leads to surface smoothing~\cite{Clarenz2000}. On the other hand, minimizing the Gaussian curvature forces the minimum curvature to be zero, resulting in flat surfaces with sharp straight edges. The visual impact on the reconstructed surfaces is shown in the supplementary material. An example of a regularized reconstruction using Gaussian curvature is shown in Fig~\ref{fig:exper_curvature}.

\section{Conclusions}

With DiffNeRF, a variant of NeRF that relies on differential geometry to regularize the depth or the normals of the learned scene, we demonstrated that it is possible to achieve state-of-the-art novel view synthesis and depth estimation in few-shot neural rendering with a simple yet flexible regularization framework. 
This is made possible by modern deep learning frameworks, which already provide the necessary tools to implement differential geometry operators, thus facilitating their use in practice.  However, the use of differential geometry is still subject to certain limitations. Higher-order operators can be costly both in memory and in computation time so a careful choice of the regularization term is essential. Operators should be chosen differently depending on the problem at hand. For example, a Gaussian curvature regularization may be appropriate for flat surfaces with strong edges, such as buildings, but could fill holes in irregular surfaces. The vast literature on differential geometry opens up many exciting opportunities to define new regularization tools with the appropriate mathematical formalism, which we hope pushes the limits of neural rendering even further.

\section{Appendix}
\label{sec:appendix}

\subsection{Additional visual results}

\begin{figure*}
    \centering
    \resizebox{\linewidth}{!}{
    \small
    \begin{tabular}{c@{\hskip 0.01\linewidth}c@{\hskip 0.01\linewidth}c@{\hskip 0.01\linewidth}c@{\hskip 0.01\linewidth}c@{}}
        \vspace{0.5em}
        & RegNeRF~\cite{niemeyer2021Regnerf} & DiffNeRF (depth. reg.) &  DiffNeRF (normals reg.) & Ground truth \\
    \rotatebox[origin=c, x=-1cm]{90}{flower} & \includegraphics[width=0.24\linewidth]{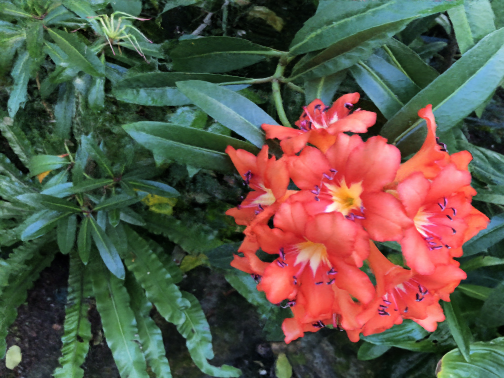} &
    \includegraphics[width=0.24\linewidth]{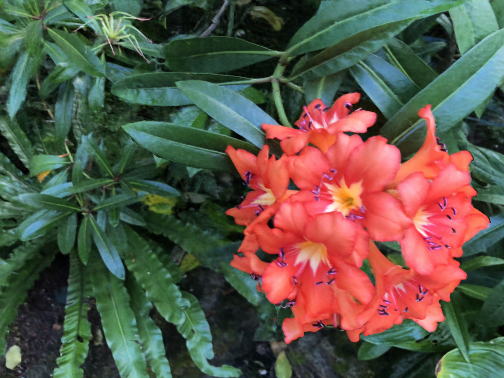} &
    \includegraphics[width=0.24\linewidth]{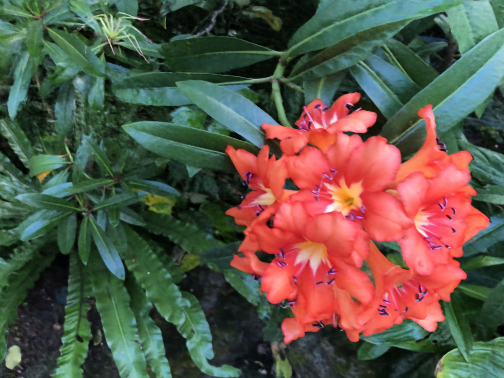} &
    \includegraphics[width=0.24\linewidth]{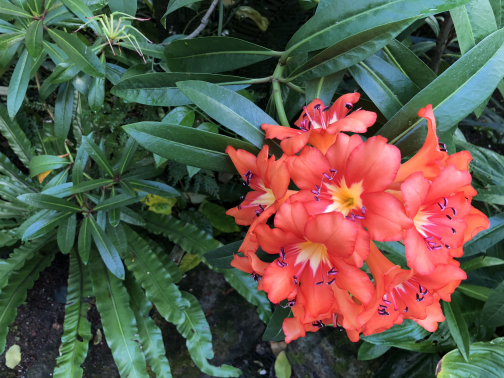} \\
    & \includegraphics[width=0.24\linewidth]{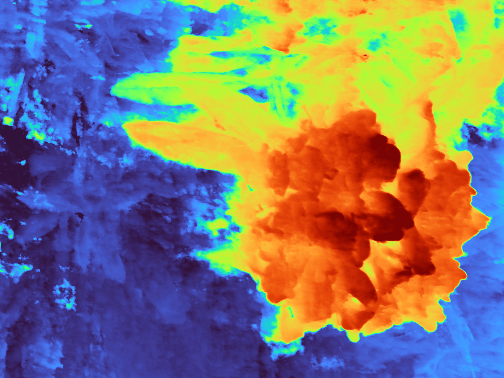} &
    \includegraphics[width=0.24\linewidth]{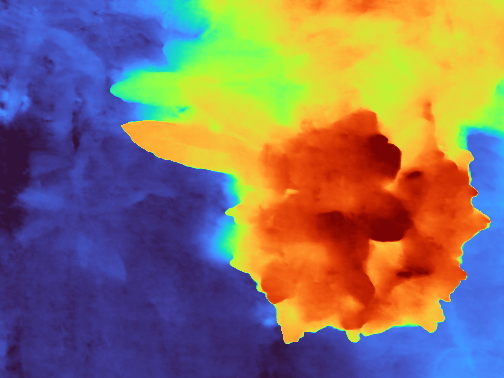} &
    \includegraphics[width=0.24\linewidth]{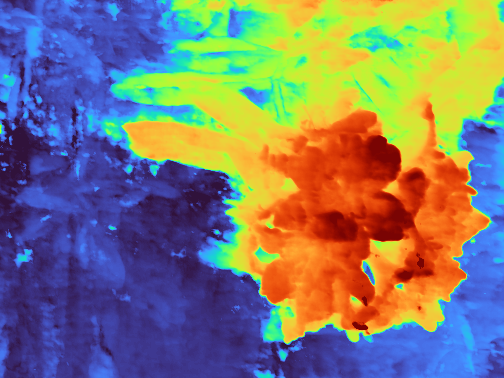} &
    \includegraphics[width=0.24\linewidth]{images/llff/blank_llff.pdf} \\
    & \includegraphics[width=0.24\linewidth]{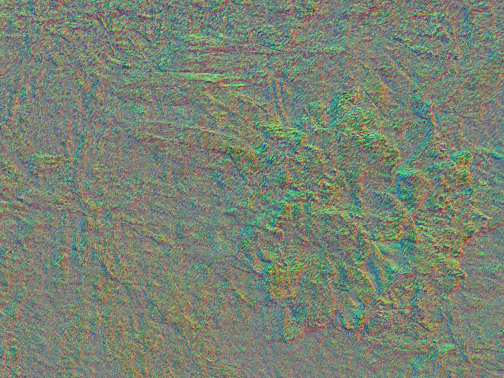} &
    \includegraphics[width=0.24\linewidth]{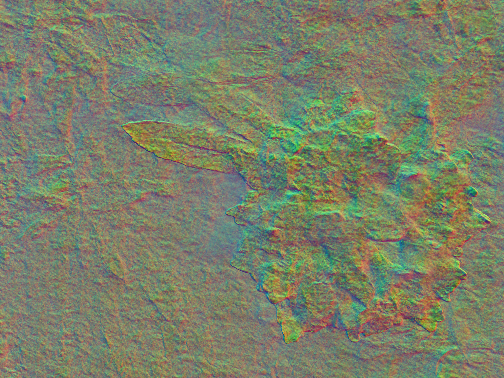} &
    \includegraphics[width=0.24\linewidth]{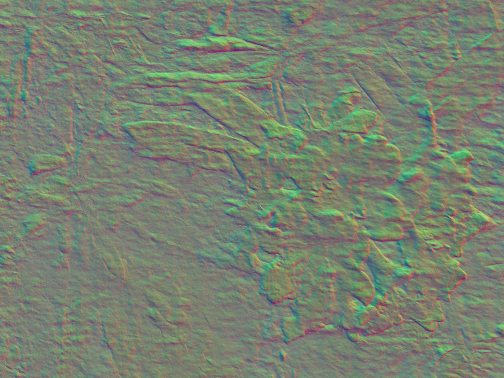} &
    \includegraphics[width=0.24\linewidth]{images/llff/blank_llff.pdf} \\
    \rotatebox[origin=c, x=-1cm]{90}{fortress} & \includegraphics[width=0.24\linewidth]{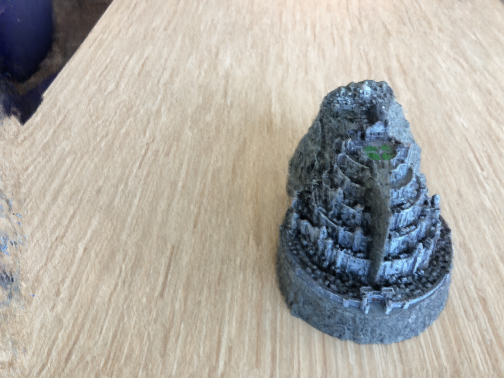} &
    \includegraphics[width=0.24\linewidth]{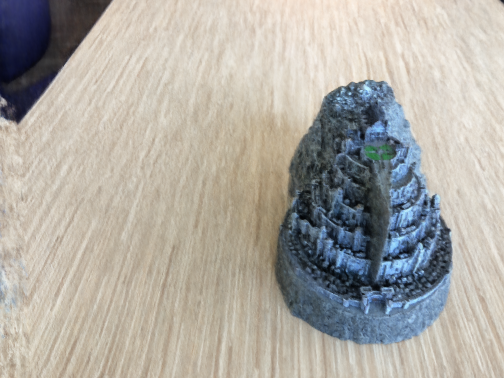} &
    \includegraphics[width=0.24\linewidth]{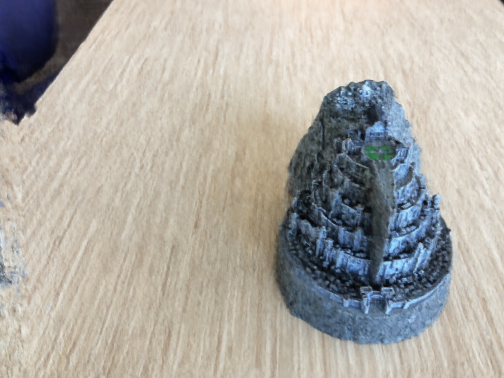} &
    \includegraphics[width=0.24\linewidth]{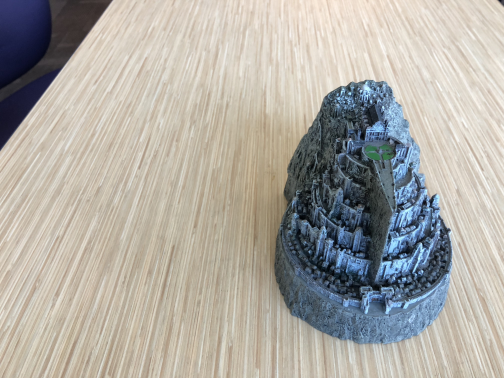} \\
    & \includegraphics[width=0.24\linewidth]{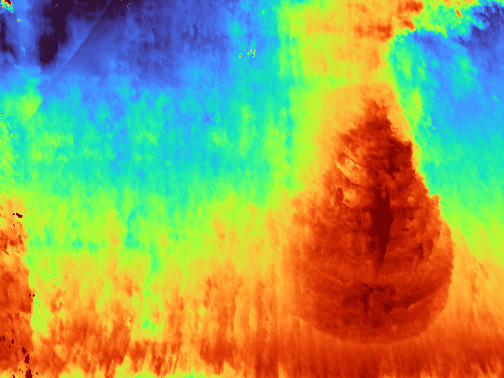} &
    \includegraphics[width=0.24\linewidth]{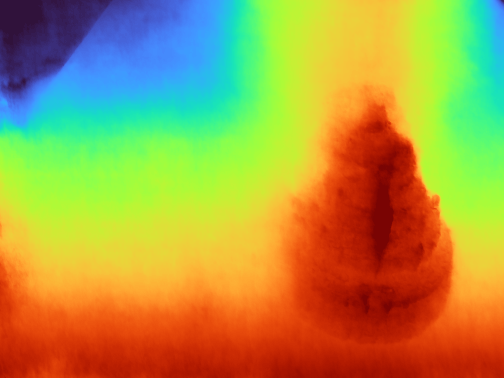} &
    \includegraphics[width=0.24\linewidth]{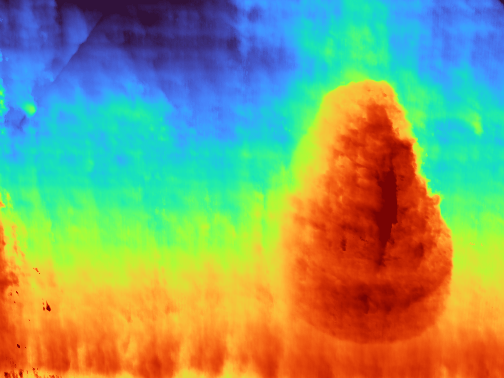} &
    \includegraphics[width=0.24\linewidth]{images/llff/blank_llff.pdf} \\
    & \includegraphics[width=0.24\linewidth]{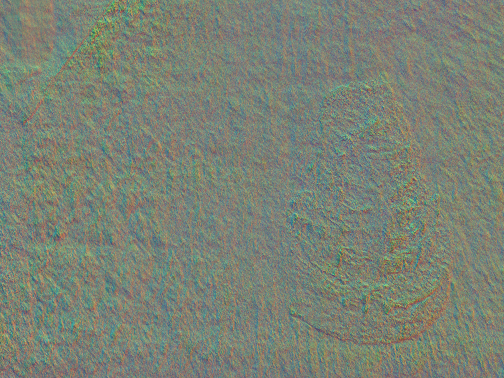} &
    \includegraphics[width=0.24\linewidth]{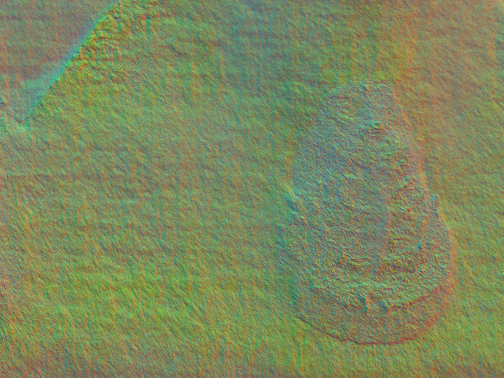} &
    \includegraphics[width=0.24\linewidth]{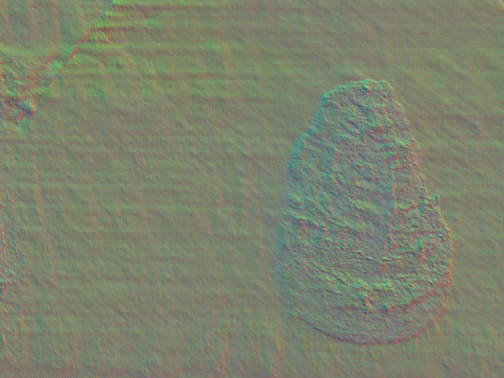} &
    \includegraphics[width=0.24\linewidth]{images/llff/blank_llff.pdf} \\
    \end{tabular}
    }
    \vspace{-1em}
    \caption{Visual examples of novel view synthesis for the \emph{flower} (top) and \emph{fortress} (bottom) sequences of the LLFF dataset after training with \ul{three views}.}
    \label{fig:visual_results}
\end{figure*}

We present additional visual results comparing RegNeRF~\cite{niemeyer2021Regnerf} to the proposed regularization framework (with both depth regularization and normals regularization) in Fig.~\ref{fig:visual_results}.

\subsection{Parameters study for surface estimation}

\begin{figure*}
    \centering
    \begin{tikzpicture}
    \node[anchor=south west, inner sep=0] (img) at (0,0){
    \begin{tabular}{c@{\hskip 0.1cm}c@{\hskip 0.1cm}c}
    \includegraphics[width=0.3\linewidth]{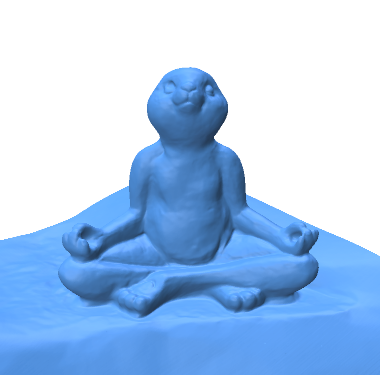}&
    \includegraphics[width=0.3\linewidth]{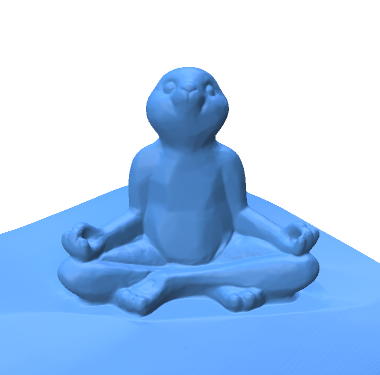}&
    \includegraphics[width=0.3\linewidth]{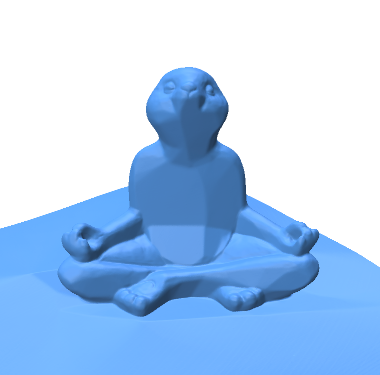}\\
    \includegraphics[width=0.3\linewidth]{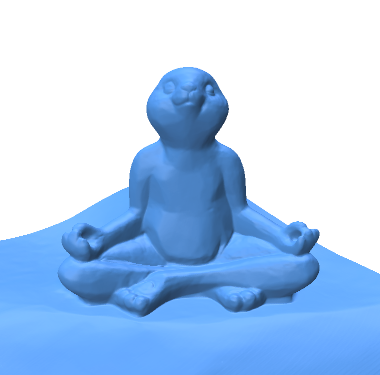}&
    \includegraphics[width=0.3\linewidth]{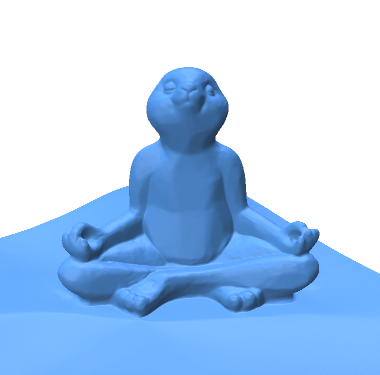}&
    \includegraphics[width=0.3\linewidth]{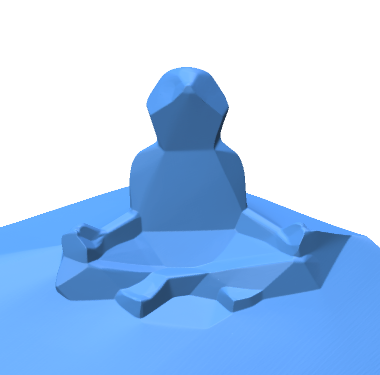}\\
    \includegraphics[width=0.3\linewidth]{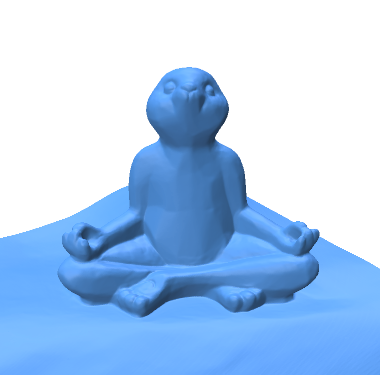}&
    \includegraphics[width=0.3\linewidth]{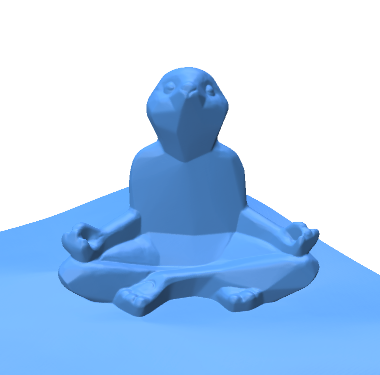}&
    \includegraphics[width=0.3\linewidth]{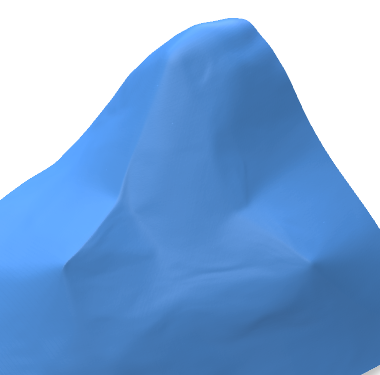}\\
    \end{tabular}};
    \begin{scope}[x={(img.south east)},y={(img.north west)}]
    \draw[-{Latex[length=2mm, width=2mm]}](-0.01, 1.00) -- (-0.01, 0.08);
    \draw[-{Latex[length=2mm, width=2mm]}](-0.03, 0.98) -- (0.92, 0.98);
    \node[text width=0.5cm]  at (-0.02,0.86) {1};
    \draw (-0.01, 0.86) -- (0., 0.86);
    \node[text width=0.5cm]  at (-0.02,0.53) {5};
    \draw (-0.01, 0.53) -- (0., 0.53);
    \node[text width=0.5cm]  at (-0.02,0.2) {10};
    \draw (-0.01, 0.2) -- (0., 0.2);
    \node[text width=1cm]  at (-0.015,0.05) {\large $\kappa_{curv}$};
    \node[text width=1cm]  at (0.17,1.0) {0.0001};
    \draw (0.17, 0.97) -- (0.17, 0.98);
    \node[text width=1cm]  at (0.5,1.0) {0.0005};
    \draw (0.5, 0.97) -- (0.5, 0.98);
    \node[text width=1cm]  at (0.83,1.0) {0.001};
    \draw (0.83, 0.97) -- (0.83, 0.98);
    \node[text width=1cm]  at (0.97,0.98) {\large $\lambda_{curv}$};
    \end{scope}
    \end{tikzpicture}
    \caption{Study of the impact of the two parameters controlling the strength of the regularization, i.e. the regularization weight $\lambda_{curv}$ and the clipping value $\kappa_{curv}$, on the scene 110 of the DTU dataset. While intermediate parameters, such as $\lambda_{curv}=0.0005$ and $\kappa_{curv}=5$, provide a good balance between detail preservation and surface smoothness, strong regularization can produce surprising surfaces such as the Cubist-like representation given by $\lambda_{curv}=0.001$ and $\kappa_{curv}=5$. These results were computed using the Gaussian curvature. Results best seen zoomed.}
    \label{fig:params_impact}
\end{figure*}

In Fig.~\ref{fig:params_impact}, we show the impact of the two parameters controlling the loss, namely the regularization weight $\lambda_{curv}$ and clipping value $\kappa_{curv}$, on a specific example of the DTU dataset~\cite{jensen2014large} using Gaussian curvature. A very small weight and a strong clipping ($\lambda_{curv}=0.0001$ and $\kappa_{curv}=1$) leads to barely no regularization as expected. On the contrary, a large weight with little clipping ($\lambda_{curv}=0.001$ and $\kappa_{curv}=10$) learns a sort of envelope of the surface. The most interesting results, however, are obtained with intermediate parameters. For example, $\lambda_{curv}=0.0005$ and $\kappa_{curv}=5$ produces a surface that is both strongly regularized while keeping most of the details. Another interesting result is that of $\lambda_{curv}=0.001$ and $\kappa_{curv}=5$, which leads to a Cubist-style surface, with little details and exaggerated edges. Such models might be interesting to learn a surface that can be represented by a mesh with few triangles or for deriving a piece-wise developable surface~\cite{sellan2020developability}.
Note also how the regularization has a tendency to fill small gaps (for example the mouth of the bunny is completely regularized with parameters $\lambda_{curv}=0.0001$ and $\kappa_{curv}=10$). This shows the importance of clipping to preserve small details and the limits of such type of regularization. This study was done using VolSDF~\cite{yariv2021volume}.

{\small
\bibliographystyle{ieee_fullname}
\bibliography{refs}
}

\end{document}